# Sensors, SLAM and Long-term Autonomy: A Review

Mubariz Zaffar
Embedded and Intelligent Systems Lab
University of Essex
mz18963@essex.ac.uk

Shoaib Ehsan
Embedded and Intelligent Systems Lab
University of Essex
sehsan@essex.ac.uk

Rustam Stolkin
Extreme Robotics Lab
University of Brimingham
r.stolkin@cs.bham.ac.uk

Klaus McDonald Maier
Embedded and Intelligent Systems Lab
University of Essex
kdm@essex.ac.uk

*Abstract*—Simultaneous Localization and Mapping, commonly known as SLAM, has been an active research area in the field of Robotics over the past three decades. For solving the SLAM problem, every robot is equipped with either a single sensor or a combination of similar/different sensors. This paper attempts to review, discuss, evaluate and compare these sensors. Keeping an eye on future, this paper also assesses the characteristics of these sensors against factors critical to the long-term autonomy challenge.

## I. INTRODUCTION

Ranging from healthcare to agriculture, mining sector to self-driving vehicles, planetary expeditions to nuclear environments, robots have found application in almost every aspect of life today [1]. For operating autonomously, a robot needs to understand its environment and should be able to localize itself within it. This problem is generally known as Simultaneous Localization and Mapping (SLAM). The core research goals in SLAM have been efficient mapping topologies [2], feature extraction and matching [3,4,5], location estimation [6,7] and loop closure techniques [8]. Interestingly, research in each of these areas has mostly been fuelled by the underlying sensor technology–the focus of this paper.

The early SLAM systems used range sensors like acoustic sensors and LIDAR [9]. These sensors provide accurate depth information but are not rich in features. The later systems mostly used vision (monocular cameras and Omni-directional cameras) as the primary source of feature-abundant information but lacked depth estimates [9]. Some of these sensors have been introduced in [10]. More recently, we have seen sensors capable of both ranging and vision in the form of RGB-D sensors and stereo-cameras [11]. After the 1st International Workshop on Event based vision at ICRA'17, a new research avenue opened for Event Cameras based Visual-SLAM Systems. This paper attempts to not only review all these sensors but also evaluate them for their deployment practicality based on their power consumption, range, price, accuracy and physical constraints. The paper also gauges these sensors against factors like lifetime, field-operability, ease-of-replacement and environmental suitability, which are critical for long-term autonomy applications.

The paper is organised as follows. Section II discusses the different types of sensors used in SLAM and their attributes. Section III assesses these sensors against factors that are critical for long-term autonomy. Finally, conclusions are presented in Section IV.

## II. SENSORS

This section discusses all the different sensors used in SLAM and their attributes. A summary of different sensor attributes is given in Table 1.

Table 1: Summary of Sensor Attributes

| Sensor Type | Power Consumption (W) | Depth Range (m) | Price ($) | Dimensions |
|---|---|---|---|---|
| Acoustic | 0.01-1 | 2-5 | 10-500 | Very Compact |
| LIDAR | 50-200 | 50-300 | 5k-100k | Bulky |
| Monocular Camera | 0.01-10 | NA | 100-5k | Very Compact |
| RGB-D | 2-5 | 3-5 | 150-400 | Compact |
| Stereo Camera | 2-15 | 5-20 | 500-5k | Compact |
| Event Camera | 0.15-1 | NA | 3k-5k | Compact |
| Omni-directional Camera | 1-20 | NA | 100-1k | Very Compact |

### A. Acoustic Sensors

Acoustic sensors have been widely used in solving the SLAM problem. An early implementation of this is [12]. In [13], the authors show an implementation of Acoustic-SLAM using moving microphone array and surrounding speakers. Assuming an Omni-directional acoustic sensor and receiver, [14] presents Echo-SLAM with a co-located microphone and acoustic source. Using landmarks as nodes of a sensor network, authors in [15] have shown a range-only SLAM system working in conjunction with sensor networks.

Acoustic sensors are basically Sound Navigation and Ranging (SONAR) sensors. They locate objects from the echo of a signal that is bounced off the object. Ultrasonic sensor, which is a sub-category of SONARs, is widely utilised for robots [16]. They use sound waves in the ultrasonic range (above 20kHz) which are not audible for human beings. These sensors report their range measurements based on the "Time-of-Flight" principle, which is the total time taken by the wave from emission to reception upon reflection.

These sensors are immune to colour and transparency, making them suitable for dark environments. Although these sensors work well under dust, dirt and high moisture content but heavy build-up of these can effect sensor's readings [18]. In addition, since the sound waves require a medium for traversal, the ultrasonic sensors do not work in vacuum. They are also effected by soft materials which tend to absorb longitudinal waves instead of reflecting them [18].

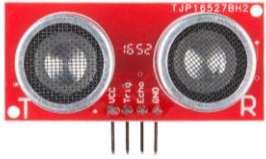

Figure 1: HC-SR04 Ultrasonic Sensor [19]

Acoustic sensors are generally very compact and come in packages of a few inch$^3$. The dimensions of HC-SRO4 (a commonly used ultrasonic sensor shown in Figure 1) are 0.814 inch$^3$ [20]. In addition, power requirement for most acoustic sensors used in robots varies from a few milli-watts to a few watts. For example, the power consumption of HC-SR04 is 75 mW [20]. They have a maximum depth range of a few meters. For most industrial applications, acoustic sensors provide an accuracy of 1% to 3% of maximum depth. For example, the depth range of HC-SR04 is 4m with an accuracy of up to 3mm in ideal conditions [21]. The accuracies and range of acoustic sensors change with environment's temperature, moisture level and air pressure for which compensation techniques are usually employed [22]. Acoustic sensors, especially ultrasonic sensors, are usually very cost-effective and can be purchased from a few USD to a few hundred USD depending upon the sensor range, depth accuracy and additional features like self-cleansing etc. For example, HC-SR04 has a current price of $1.5.

### B. LIDAR

LIDAR has driven a lot of research in range-based SLAM systems. An early work utilizing LIDAR in conjunction with Rao-Blackwellized particle filters is presented in [23]. Authors in [24] present an approach for finding interest regions in the data coming from Laser Sensors. Using occupancy grid-maps for mapping, [25] shows a scalable SLAM system with full-estimation of 3D pose. In [26], real-time loop-closure is achieved with a LIDAR.

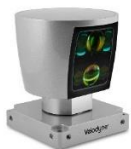

Figure 2: Velodyne HDL-64E [27]

LIDAR is an acronym for Light Detection and Ranging. The underlying technology for LIDAR has been around since 1960s. The working of LIDAR is like acoustic sensors. Instead of using sound waves, LIDAR employs electromagnetic waves (light) as a mean of radiation. By firing up to 1,000,000 pulses per second a LIDAR generates a 3D Visualization of its surroundings known as the 'Point Cloud'. LIDARs can provide 360 degrees of visibility and are very accurate (~ ±2 cm) in calculating the depth information. A commonly used LiDAR for self-driving cars is the Velodyne HDL-64E (shown in Figure 2).

LIDARs are usually bulky sensors but light-weight compact variants have been introduced recently as well. For example, HDL-64E weighs around 13Kg and has dimensions of 8 inches × 8 inches × 11 inches. However, VLP-16 and HDL-32 are much lighter and compact. In general, LIDARs are usually power-hungry sensors. The power usually ranges between a few watts to a few hundred watts depending upon the sensor range and features. For example, the power requirement of Velodyne HDL-64E is 80 Watts [28]. LIDARs have the most depth range of all the sensors employed for SLAM. This range usually varies from 50 meters to a few hundred meters depending upon the manufacturer and selected product. They are also accurate with an angular resolution of 0.1 to 1 degree and depth accuracy of a few centimetres. For example, Velodyne HDL-64E has a range of 120 m and a depth accuracy of ±2 cm. These sensors are usually expensive. The high-end HDL-64E costs about $75,000 whereas VLP-16 costs about $8,000 [29]. Recently, price of VLP-16 was cut by half [30].

### C. Monocular Camera

A popular sensor used for SLAM is a monocular camera. For example, a real-time SLAM system based on a single camera is presented in [31] and [32]. A SLAM system resulting from the fusion of monocular cameras and inertial sensors is proposed in [33]. A comparison of monocular SLAM and Stereo SLAM is presented by authors in [34]. A semantic SLAM system for a monocular camera is proposed in [35].

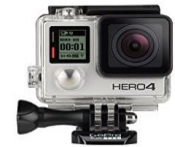

Figure 3: GoPro Hero-4 [38]

Monocular cameras are standard RGB cameras. A GoPro camera Hero-4 is shown in Figure 3. One of the main reasons for the usage of pure monocular SLAM is simple hardware [36]. For example, monocular SLAM is possible on mobile phones without the need for additional hardware. On the contrary, the algorithms needed for monocular SLAM are much more complex due to the inability to directly infer the depth information from a static image.

Cameras are usually very compact and a typical camera has dimensions between 5 inch$^3$ to 20 inch$^3$. Since Cameras are passive sensors, their power consumption is low. The usual range of power consumption is within a few watts. A typical GoPro camera consumes power between few milli-watts to ~5 watts depending on the resolution and mode settings [37]. There is an enormous amount of variety available for the selection of monocular cameras. Depending on the resolution, visual features, mounting, storage, power source, a camera costs from a few dollars to a few thousand dollars. For example, the GoPro Hero Series prices vary from $150 to $450 depending upon the model.

### D. RGB-D Sensors

As RGB-D sensors possess the strengths of both Range-based and Vision-based sensors, they draw a lot of interest from the robotics community. Authors in [39, 40, 41] present an evaluation of RGB-D SLAM. A real-time, large-scale dense SLAM system is developed in [42] using RGB-D sensors. The application of RGB-D SLAM to aerial systems is shown in [43].

RGB-D sensors are a combination of monocular camera, IR transmitters and IR receivers. They provide RGB detail of a scene along with the estimated depth of each pixel. Microsoft, in November 2010, released the Kinect RGB-D sensor shown in Figure 4. The Kinect, like other RGB-D sensors, provides colour information as well as the estimated depth for each pixel but is relatively inexpensive.

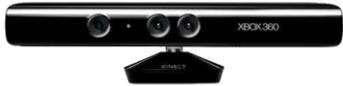

Figure 4: Kinect RGB-D Sensor [44]

There are two depth calculation techniques used in RGB-D sensors, Structured Light (SL) and Time-of-Flight (TOF). The SL technique, as in PrimeSense sensor, projects an infrared speckle pattern. The projected pattern is then captured by an infrared camera in the sensor, and compared part-by-part to reference patterns stored in the device. These patterns were captured previously at known depths. The sensor then estimates the per-pixel depth based on with which reference patterns the projected pattern matches [45]. The TOF technique is similar to that employed in LIDAR. The first version of Microsoft Kinect used SL as a depth calculation mechanism while the recent versions of Kinect use TOF.

The original Kinect consumed power around 2.5 Watts, while the most recent Kinect Version 2 consumes about 15 Watts. Power consumption for other RGB-D sensors like Asus Xtion also lies within this range [46]. There is an inverse relationship between the range and depth accuracy for RGB-D sensors. Kinect V1 and Asus Xtion have a maximum range of 3.5m, while the Kinect V2 has a range of 4.5m [46]. The price range for RGB-D sensors is $150 to $400 depending on manufacturer and specifications. RGB-D sensors are relatively compact but have a wide base. The dimensions for Kinect, Xtion and others are usually ~10 inches × ~2.6 inches × ~2.6 inches.

### E. Stereo Cameras

Stereo cameras have also driven a lot of research in the area of visual SLAM. Authors in [47], present an implementation of Stereo-SLAM using particle filters. To perform SLAM in large indoor and outdoor environments, [48] presents a 6-dof SLAM using hand-held stereo camera. Using iterative closest point algorithm, [49] shows a robust 3D stereo camera SLAM. In [50], the ability of stereo cameras to provide depth information in addition to the conventional multi-view disparity-based depth calculation is exploited

Inspired from the human eye, stereo cameras use the disparity of two camera images looking at the same scene to calculate depth information. Unlike RGB-D cameras, stereo cameras are passive cameras. They do not suffer from the problem of scale-drift as found in monocular cameras [51]. Some popular stereo-cameras are Bumblebee 2 (shown in Figure 5), Bumblebee XB3, Surveyor Stereo Vision System (SVS), Ensenso, Capella and Minoru 3D Webcam.

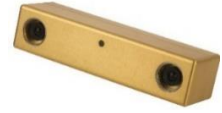

Figure 5: Bumblebee Stereo Camera [52]

The power requirements for most stereo cameras range between 2 to 15 Watts. For example, the power requirement of Bumblebee 2 and Bumblebee XB3 are 2.5 Watts and 4 Watts respectively [53]. The typical maximum range of stereo cameras is between 5 to 20 meters at varying depth resolution. The depth accuracy usually varies from a few millimetres to ~5 centimeters at maximum depth. Depth range and accuracy are linked in [54]. The cost of stereo cameras varies from a few hundred USD to a few thousand USD depending on the camera resolution, range and other specifications. For example, the price of Bumblebee XB3 is $3500. The physical dimensions of stereo cameras usually depend on the baseline. Most cameras can be found in dimensions under 6 inches × 2 inches × 2 inches.

### F. Event Cameras

Since the intrinsic nature of Event cameras is different from traditional cameras, a separate SLAM paradigm has opened-up. Parallel Tracking and Mapping (PTAM) is one of the major SLAM techniques and [55] shows an implementation of PTAM for event cameras. Although event cameras are an excellent choice for dynamic scenes but in static scenes, they give little-to-no information. Thus, [56] combines event cameras and monocular cameras for an Ultimate SLAM. Authors in [57] present a complete continuous-time event-based SLAM system in conjunction with inertial measurements. An event-camera training dataset can be accessed through [58].

Event cameras, such as the Dynamic Vision Sensor (DVS128 in Figure 6), are bio-inspired vision sensors that output pixel-level brightness changes instead of standard intensity frames [59]. They offer significant advantages over standard cameras, including a very high dynamic range, no motion blur, and a latency in the order of microseconds [60]. Since, their output is composed of a sequence of asynchronous events rather than actual intensity images, traditional vision algorithms are not exactly applicable. So, new algorithms that exploit the high temporal resolution and asynchronous nature of the sensors have been proposed and researched.

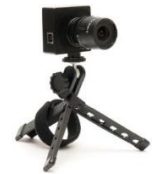

Figure 6: DVS128 Event Camera [61]

Since event cameras by default don't capture redundant information in scenes, they are very power-efficient. For example, the power consumption for DVS Cameras varies from 150mW to 1 W [62]. Like monocular cameras, event cameras also do not provide any depth information, so the concept of range is not exactly applicable. However considerable work has been done on depth estimation and 3D reconstruction with multiple temporal views of similar objects using event cameras [63]. Event cameras at present are expensive and range between $3,000 to $5,000 depending on the latency, resolution and other specifications. These cameras usually come in compact packages. The smallest version from iniVation is the 0.7inches × 0.7inches × 0.3inches mini-eDVS, while the bulkiest is 3 inches × 2 inches × 1 inch eDVS.

*G. Omni-directional Cameras*

An early implementation of SLAM with an Omni-directional camera can be observed in [64]. Authors in [65] combine particle filters with a SIFT feature extractor for images obtained from Omni-directional camera. An extensive review of SLAM based on Omni-directional camera is presented in [66].

Omni-directional cameras are RGB cameras with 360 degrees field of view. In contrast to traditional computationally expensive stitching of images in panoramic photography, an omnidirectional camera can be used to create panoramic art in real-time, without the need for post processing, and typically gives much better quality. A Samsung Gear 360 utilizing two fish-eye lenses is shown in Figure 7.

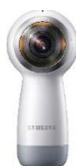

Figure 7: Samsung Gear 360 [67]

Commercially, there are many variants of Omni-directional cameras. The parameters discussed here are based on some of the popular 360 cameras like GoPro Fusion, Ricoh Theta V, Detu Twin, Samsung Gear 360, LGR105 and Yi Technology 360VR. Omni-directional cameras are highly compact. Their maximum dimensions are usually up to 4 inches × 2 inches × 2 inches. The power requirement for 360 cameras usually varies from 1 to 20 Watts [68]. These cameras do not provide any depth information. Due to their wide field-of-view, they can easily capture objects in multiple directions which makes them very suitable for Visual-SLAM systems. The pixel resolution is unmatched and most provide 4K images and videos. Depending on the manufacturer and camera features, prices vary from $100 to $1000. For example, Gear-360 costs $219.

## III. EVALUATION AGAINST LONG-TERM AUTONOMY

This section assesses the sensors mentioned in Section II against factors, such as sensor life time, field operability, ease of replacement, and environmental suitability, which are critical to the long-term autonomy challenge.

*A. Acoustic Sensors*

In terms of sensor life time, acoustic sensors have shown excellent performance in the industry over the last few decades. For ceramic diaphragm-based sensors, performance decays logarithmically. i.e., the decay in performance over the first 10 days is the same as the decay over the next 100 and subsequently 1000 days [69]. Since ultrasonic sensors are usually used in industrial environments, their designs are rigid and show acceptable performance in varying weather conditions. Their working temperature range is -20 to +80 degrees Celsius for most sensors. Some are also water-proof and designed to withstand harsh environments; for example, ToughSonic line from Senix is IP68 / NEMA-4X / NEMA-6P rated [70]. However, noise caused by air nozzles, pneumatic valves or solenoids, and ultrasonic welders do affect the performance of these sensors.

Regarding ease of replacement; due to their compactness and minimal pinout, they are usually plug-and-play and can be deployed as hot-swappable sensors. As per the UK's independent advisory group's report on Non-Ionizing radiation 2010, the exposure limit for Sound Pressure Level on general public is set to 70 dB (at 20 kHz), and 100 dB (at 25 kHz and above). The SPL for ultrasonic sensors operating above 25KHz is in the safe range of ~100dB and follows the inverse distance (1/r) law [71]. However constructive interference of ultrasonic transmitters in multiple co-located systems may pose a challenge and needs to be investigated.

*B. LIDAR*

Due to the rotary nature of LIDAR, it is expected to suffer from wear-and-tear over a period of time. New technologies involving electronic handling of rotation instead of mechanical parts are also promising [72]. As the main target audience for LIDAR has been driverless cars and autonomous robots, their designs are rugged and suitable for harsh environments. The usual working temperature range is -50 to +80 degrees Celsius and they come in rugged housings [73]. However, owing to the dynamic nature of some variants of LIDAR, frequent servicing may be required. Due to bulkiness and heaviness, LIDARs need to be handled with care when replacing them. Lasers used in LIDAR are Class 1 Eye-Safe per IEC 6–825–1: 2007 & 2014. However, the risks of multiple lasers in constructive interference is yet to be analysed. Unlike cameras, they generally do not raise any privacy concerns.

*C. Monocular Camera*

The life time of cameras is usually calculated based on its shutter life. i.e., the number of shutter actuations before which the camera will go blind. Most cameras have a shutter life between 30,000 to 150,000 clicks [74]. Digital video-cameras, on the other hand, have a combination of mechanical and electronic shutters. The mechanical shutter actuation is performed at the start and end of capturing a video sequence. Although electronic shutters do not have any mechanical movement (thus no shutter life based on mechanical wear-and-tear), however, long exposure to light can damage image sensors which is why additional mechanical shutter actuations are needed. While today's shutter life appears very attractive for professional photography, evaluation for long-term vision-based SLAM systems needs to be performed.

Cameras have been widely used in field before, and some show excellent performance. Most adventure and monitoring cameras are water-proof, heat-resistant and resilient. Due to their compactness, light-weight, serial communication lines, and ease of mounting; they are easy to replace. Since Cameras are only passive elements they don't pose any threat to the environment other than their contribution to e-waste like any other electronics. However, wide-spread use of cameras on mobile robots can raise privacy concerns which needs to be addressed.

*D. RGB-D Sensors*

In terms of life time, most RGB-D sensors have shown promising performance over the years. However, rigorous use in extreme environments is an area where Kinect and other RGB-D sensors need to be tested for longer periods. Primarily, RGB-D sensors were developed as user-interfaces and not for mobile platforms in real world. Although the intrinsic nature of these sensors is the same as cameras, their compatibility with harsh environments needs to be investigated. Due to their compactness and serial interface, they are easy to replace in field.

Regarding environmental suitability, the IR laser in Kinect is 780nm (visible light ending at 760 nm, thus making the Kinect laser very short wave IR, or SWIR). Furthermore, in order to cover a wide area the radiation pattern is diffused. It is rated as a Class 1 laser device [75] and complies with IEC 60825-1: 2007-3. However, wide spread usage of RGB-D cameras, may raise the same privacy concerns as for standard cameras.

*E. Stereo Camera*

The underlying hardware for stereo cameras is the same as monocular cameras and thus they also show acceptable performance over the years depending upon shutter life. However, evaluation for long-term vision-based SLAM needs to be performed. Stereo cameras have been used in fields on various static and dynamic platforms. Recently many versions of drones utilized stereo cameras as well. However, since the performance of stereo cameras is highly dependent upon environment and scenes, their multi-terrain usefulness needs to be investigated. Most Stereo cameras come with different mounting styles and have serial interface. Due to their compactness, they are easy to replace. Stereo cameras, being passive sensors, do not emit any sort of energy. They do contribute to electronic waste and pose challenges of privacy.

*F. Event Cameras*

Since event cameras have recently emerged, their lifetime is yet to be evaluated. However, based on our correspondence with manufacturers, they have tested these cameras and expect these to show excellent performance over a period of 6-7 years. The DVS event cameras are designed specifically for the task of robot navigation. They have strong mounting, low power-consumption but evaluation in extreme weathers should be performed. These cameras are compact, serially interfaced and have different mounting options, making them easy-to-replace. These are passive sensors and do not rely on any emissions for mapping the environment.

*G. Omni-directional Cameras*

Due to their profound use in tourism, these cameras are designed to last for years, similar to other standard cameras based on their shutter life. However, that evaluation is for occasional image/video capturing and not applicable to long-term autonomous robots. Most of the 360 cameras are designed for rugged use in harsh environments and some are water-proof. This provides great benefits for field operations. Due to their highly compact sizes and serial interface they can be easily replaced. Being passive sensors, they do not pose any radiation threats. However, like standard cameras they can also raise privacy concerns when used in conjunction with autonomous robots.

IV. CONCLUSIONS

It can be concluded that research in the area of SLAM is highly dependent on the sensing technologies and much innovation is yet to come. Acoustic sensors lack feature rich representation of environment. LIDARs are expensive, bulky and pose computational challenges. Monocular, omni-directional and event cameras lack depth information. Additionally, event cameras fail in static scenes. Although RGB-D cameras provide depth information, they have limited range. A sensor with the dynamic response of event cameras, static feature-rich scans of omni-directional cameras and range-prowess of LIDARs, may achieve the ultimate SLAM system. And yet after that, it needs to be reasonably priced, power-efficient, compact, lifelong and environment-friendly to realize long-term autonomy.

ACKNOWLEDGMENT

This work is supported by the UK EPSRC through grants EP/R02572X/1 and EP/P017487/1.